\documentclass[10pt,twocolumn,letterpaper]{article}

\usepackage[pagenumbers]{wacv} %

\usepackage{adjustbox}
\usepackage{graphicx}
\usepackage{colortbl}
\usepackage[table]{xcolor}
\usepackage{multirow}

\usepackage{dblfloatfix}

\usepackage{booktabs}
\usepackage{siunitx}  
\usepackage{multirow} 
\usepackage{graphicx} 

\usepackage{subcaption}

\newcommand{\BGcolor}[3][HTML]{\definecolor{mycolor}{HTML}{#2}\bgcolor{mycolor}{#3}}

\newcommand{\bgcolor}[2]{\setlength{\fboxsep}{0pt}\colorbox{#1}{\strut #2}}

\usepackage{balance}

\definecolor{wacvblue}{rgb}{0.21,0.49,0.74}
\usepackage[pagebackref,breaklinks,colorlinks,allcolors=wacvblue]{hyperref}

\title{OracleGS: Grounding Generative Priors for Sparse-View Gaussian Splatting}

\author{
    Atakan Topaloğlu$^{1,2,3}$ \footnotemark[1]  \hspace{1.5em}
    Kunyi Li$^{4,6}$ \hspace{1.5em}
    Michael Niemeyer$^{5}$ \hspace{1.5em}
    Nassir Navab$^{4,6}$ \\
    A. Murat Tekalp$^{2,3}$ \hspace{1.5em}
    Federico Tombari$^{4,5,6}$
    \\[8pt] %
    $^1$ETH Zürich \quad
    $^2$Ko\c{c} University \quad
    $^3$KUIS AI Center \quad
    $^4$Technical University of Munich \\
    $^5$Google \quad
    $^6$Munich Center for Machine Learning
}

\begin{document}

\twocolumn[{%
    \renewcommand\twocolumn[1][]{#1}%
    \maketitle
    \thispagestyle{empty}
    \vspace{-10mm} %
    \begin{center}
        \includegraphics[width=1.0\linewidth]{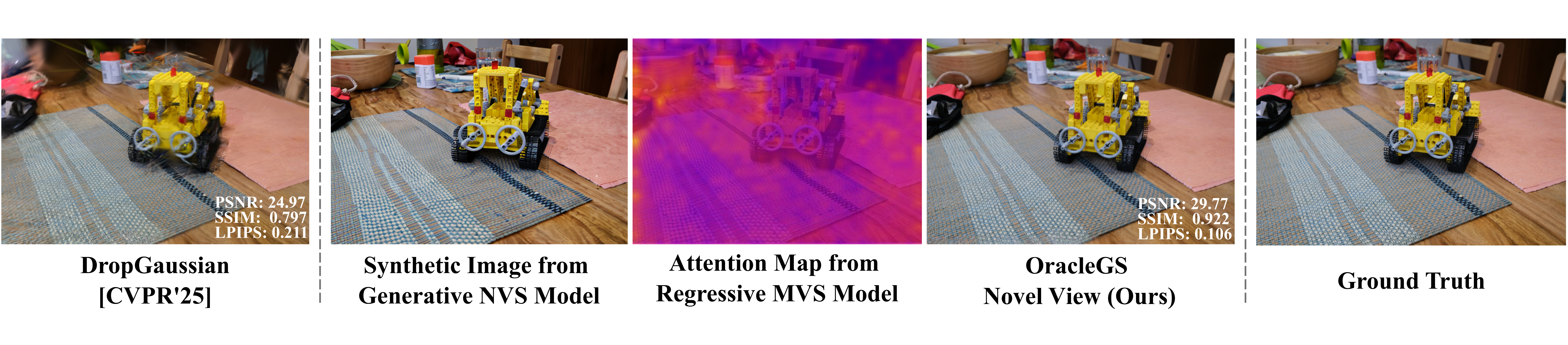}
        \vspace{-6mm} %
        \captionof{figure}{
         \textbf{OracleGS Reconciles Generative Completeness with Regressive Fidelity.} First, a generative model proposes a potentially flawed, synthetic view (middle left). Our MVS-based oracle then grounds this proposal by quantifying a 3D uncertainty map (middle), effectively identifying various sources of generative errors; including faulty textures on the mat, inconsistent structures on the lego, and under-observed backgrounds, as regions of high uncertainty (highlighted in blue). Using this signal to guide a confidence-weighted optimization, OracleGS filters these artifacts, producing novel views (middle right) with superior fidelity compared to both the synthetic input and prior state-of-the-art DropGaussian~\cite{park2025dropgaussian} (left).}
        \label{fig:teaser}
    \end{center}
}]

\footnotetext[1]{Work conducted as part of a research collaboration with Google and the Technical University of Munich (TUM).}

\begin{figure*}[t]
    \centering
    \includegraphics[width=\textwidth]{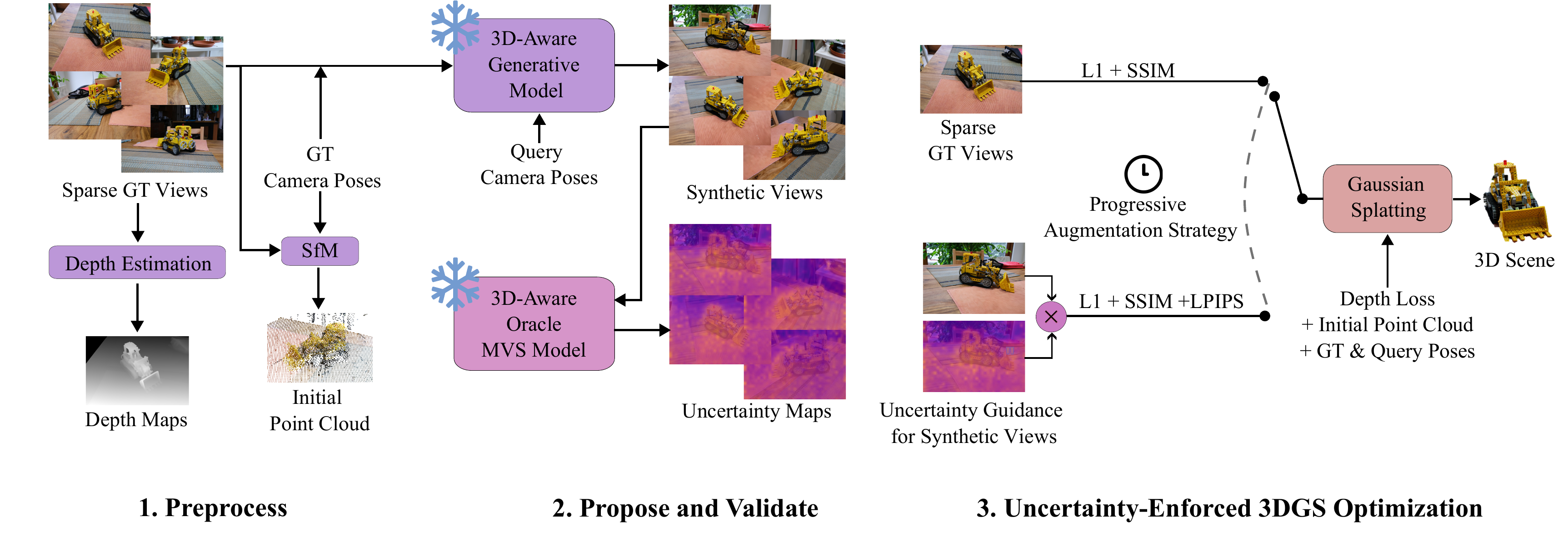} 
    \caption{\textbf{Overview}. Given sparse input views with poses, we first estimate initial point cloud and depth maps. 
    Afterwards, a 3D-Aware generative model proposes novel synthetic views, while the 3D-Aware Oracle's attention maps are used as a proxy for 3D uncertainty. 
    Finally, we train the 3DGS model using a standard loss on the GT views and our novel uncertainty-guided loss on the synthetic views. We employ a progressive augmentation strategy over the course of the optimization to control the ratio of GT and synthetic images at each iteration, which helps to stabilize training and guide scene structure.}
    \label{fig:pipeline}
\end{figure*}

\begin{abstract}
Sparse-view novel view synthesis is fundamentally ill-posed due to severe geometric ambiguity. Current methods are caught in a trade-off: regressive models are geometrically faithful but incomplete, whereas generative models can complete scenes but often introduce structural inconsistencies. We propose OracleGS, a novel framework that reconciles generative completeness with regressive fidelity for sparse view Gaussian Splatting. 
Instead of using generative models to patch incomplete reconstructions, our "propose-and-validate" framework first leverages a pre-trained 3D-aware diffusion model to synthesize novel views to propose a complete scene. We then repurpose a multi-view stereo (MVS) model as a 3D-aware oracle to validate the 3D uncertainties of generated views, using its attention maps to reveal regions where the generated views are well-supported by multi-view evidence versus where they fall into regions of high uncertainty due to occlusion, lack of texture, or direct inconsistency. This uncertainty signal directly guides the optimization of a 3D Gaussian Splatting model via an uncertainty-weighted loss. 
Our approach conditions the powerful generative prior on multi-view geometric evidence, filtering hallucinatory artifacts while preserving plausible completions in under-constrained regions, outperforming state-of-the-art methods on datasets including Mip-NeRF 360 and NeRF Synthetic.

\end{abstract}

\section{Introduction}
\label{sec:intro}

Novel-view synthesis (NVS)~\cite{mildenhall2020nerf} constitutes a fundamental problem in computer vision, which seeks to reconstruct the 3D scenes from given viewpoints and render photo-realistic observations from novel perspectives. This capability is central to a variety of downstream applications, including large-scale 3D content generation~\cite{liu2024citygaussian, lin2024vastgaussian, jiang2025horizon, Zhou_2024_CVPR}, digital twin construction~\cite{guo2025articulatedgsselfsuperviseddigitaltwin, Shimomura_2025}, and immersive virtual reality~\cite{franke2025vrsplatting, jiang2024vr-gs}. A particularly challenging yet practically significant setting is sparse-view NVS, where only a limited number of casually captured images are available acquired via handheld devices in uncontrolled environments. Addressing this setting is critical for enabling scalable and accessible 3D content creation in real-world scenarios.

However, when only a few input views are available, the underlying 3D scene structure becomes severely under-constrained, hindering high-quality synthesis. This transforms the task from a well-posed interpolation problem into a highly ill-posed extrapolation problem, where vast unobserved regions lead to significant structural ambiguity. This is particularly evident for explicit representations like 3D Gaussian Splatting (3DGS)~\cite{kerbl3Dgaussians}, which, despite offering high quality real-time rendering, are prone to severe artifacts when optimized with sparse data~\cite{xiong2023sparsegs, zhang2024cor}. Consequently, a significant body of recent work has focused on regularizing the 3DGS optimization process using techniques like generating pseudo-views~\cite{zhu2023FSGS}, applying dropout-based strategies~\cite{park2025dropgaussian}, or incorporating external priors from pre-trained models~\cite{xiong2023sparsegs, zhang2024cor} in the sparse view regime.

To address the challenge, recent research has diverged into two main paradigms, each with fundamental limitations. On the one hand, feedforward regression-based models~\cite{yu2021pixelnerf, charatan23pixelsplat, chen2024mvsplat, ziwen2025llrm, jin2025lvsm} are effective in extracting information from visible regions, but their performance degrades in regions of high 3D uncertainty, where robust correspondences cannot be established. This leads to poorly estimated surfaces and textures in occluded or under-observed parts of the scene. On the other hand, generative models, particularly diffusion-based approaches~\cite{liu2023zero1to3,
 zeronvs, 
 wu2023reconfusion, 
 gao2024cat3d, 
 VoletiYBLPTLRJ24, 
 wang2024motionctrl, 
 yu2024viewcrafter, 
watson2025controllingspacetimediffusion, zhou2025stable} excel in hallucinating plausible content for these unobserved regions. However, they often lack strict 3D consistency with the input views, producing results that are visually compelling but structurally incorrect. This leaves a critical gap: regressive methods are geometrically faithful but incomplete, while generative methods are complete but often unfaithful.

In this paper, we resolve this conflict by inverting the conventional paradigm. Instead of using a regressive model to form an incomplete reconstruction and a generative prior to fill the gaps, our "propose-and-validate" framework first leverages a powerful 3D-aware generative model to \emph{propose} a complete and plausible scene by synthesizing a dense set of novel views. We then repurpose a classical multi-view stereo (MVS) model not as a reconstructor, but as a 3D-Aware oracle that assesses where the generated views are well-supported by multi-view evidence, thereby \emph{validating} these proposals. It distills this assessment into per-pixel uncertainty maps, which directly guide the optimization of a 3DGS~\cite{kerbl3Dgaussians} representation via a uncertainty-weighted loss. This approach grounds the powerful generative prior on multi-view evidence, filtering out inconsistent artifacts while preserving plausible completions, leading to novel views that are both complete and structurally sound.
In summary, our main contributions are:
\begin{itemize}
    \item \textbf{A "propose-and-validate" framework for sparse-view NVS.} Our method, OracleGS, synergizes generative priors with geometric validation by using a 3D-aware diffusion model to propose novel views and a multi-view stereo model to validate their consistency, achieving state-of-the-art results in sparse-view-3DGS.
    \item \textbf{A novel uncertainty-weighted loss derived from MVS attention.} We introduce a novel framework that repurposes a multi-view stereo (MVS) model as a 3D uncertainty oracle, using its attention maps to formulate a new uncertainty-weighted loss for guiding generative novel view synthesis.
    \item \textbf{A progressive augmentation schedule to stabilize optimization.} We introduce a training curriculum that treats synthetic views as a temporary structural scaffold. By  modulating their influence over the course of the training, we first build a coherent global structure before yielding to ground-truth data for high-frequency refinement.
\end{itemize}

\section{Related Work}

\subsection{3D Scene Reconstruction}

\paragraph*{Neural Radiance Fields.} Neural Radiance Fields (NeRF)~\cite{mildenhall2020nerf}, represents a 3D scene as a continuous, implicit function learned by a neural network that maps 3D coordinates and viewing directions to color and density. This approach led to a series of extensions that improved rendering quality with anti-aliasing~\cite{multinerf2022}, handling complex unbounded scenes~\cite{barron2022mipnerf360}, and accelerating training~\cite{barron2023zipnerf} or improving performance~\cite{yu_and_fridovichkeil2021plenoxels, M_ller_2022}. Despite their high fidelity, NeRF-based methods typically suffer from slow training and rendering times, limiting their practical applicability.
\paragraph*{3D Gaussian Splatting.} To address these efficiency bottlenecks, 3D Gaussian Splatting (3DGS)~\cite{kerbl3Dgaussians} was introduced. Instead of a neural network, 3DGS models a scene explicitly with a set of 3D Gaussian primitives that are optimized via a differentiable rasterizer, enabling real-time rendering and significantly faster training while achieving state-of-the-art quality, which has led to their rapid adoption for novel view synthesis, including improved rendering quality~\cite{kheradmand20243d, Yu2023MipSplatting}, implicit surface reconstruction~\cite{li2025monogsdfexploringmonoculargeometric}
dynamic scene modeling~\cite{Wu_2024_CVPR}, text-to-3D generation~\cite{tang2023dreamgaussian, yi2023gaussiandreamer} while hybrid methods~\cite{niemeyer2025radsplat} use radiance fields as a prior to improve the robustness of the Gaussian optimization process.

While both NeRF and 3DGS have demonstrated success in the dense-view regime, their performance degrades significantly when the number of training images are reduced, under sparse-view conditions. This vulnerability highlights the need for strong priors to regularize the reconstruction process, a challenge our work directly addresses.

\begin{table*}[h!]
\scriptsize
\centering
\adjustbox{max width=\textwidth}{%
\centering
\fontsize{9.5pt}{11pt}\selectfont
\renewcommand{\arraystretch}{1.1} %
\begin{tabular}{c l|  c c c | c c c} %
\toprule
\multicolumn{2}{c}{\multirow{2}{*}[-0.5ex]{Methods}} & \multicolumn{3}{c}{12-view} & \multicolumn{3}{c}{24-view}\\ \cmidrule(lr){3-5}\cmidrule(lr){6-8}
\multicolumn{2}{c}{} & \multicolumn{1}{c}{PSNR($\uparrow$)} & \multicolumn{1}{c}{SSIM($\uparrow$)} & \multicolumn{1}{c}{LPIPS ($\downarrow$)} & \multicolumn{1}{c}{PSNR($\uparrow$)} & \multicolumn{1}{c}{SSIM($\uparrow$)} & \multicolumn{1}{c}{LPIPS ($\downarrow$)}\\ \midrule

& Mip-NeRF 360~\cite{barron2022mipnerf360} & 17.73 & 0.432 &  0.520  & 19.78 & 0.530  & 0.431\\

& RegNeRF~\cite{Niemeyer2021Regnerf} & 18.84 &  0.437 &  0.544& 20.55 & 0.546 & 0.398 \\
& SparseNeRF~\cite{wang2022sparsenerf} & 17.44 & 0.395 & 0.609& 21.13  & 0.600 & 0.389\\

\midrule

& 3DGS~\cite{kerbl3Dgaussians} & 18.52 & 0.523 & 0.415 & 22.80 & 0.708 & 0.276\\
& DNGaussian~\cite{li2024dngaussian} & 16.28 & 0.432 & 0.549 & 19.26 & 0.550 & 0.440\\
& FSGS~\cite{zhu2023FSGS} & 18.80 &  0.531 & 0.418 & 22.82  & 0.693 & 0.293 \\

& SparseGS~\cite{xiong2023sparsegs} & 19.37 &  0.577 & 0.398\cellcolor{yellow!35} & 23.02 & 0.713 & 0.290\\

& CoR-GS~\cite{zhang2024cor} & 19.52\cellcolor{yellow!35} & 0.558 & 0.418 & 23.39 & 0.727\cellcolor{yellow!35} & 0.271\cellcolor{yellow!35}  \\

& CoMapGS~\cite{Jang_2025_CVPR} & 19.68\cellcolor{orange!35} & 0.591\cellcolor{orange!35}  & 0.394\cellcolor{orange!35} & 23.46\cellcolor{orange!35}  & 0.734\cellcolor{orange!35} & 0.264\cellcolor{orange!35} \\

&DropGaussian (Our Replication)~\cite{park2025dropgaussian}  & 19.38 & 0.583\cellcolor{yellow!35} & 0.402 & 23.44\cellcolor{yellow!35} & 0.736\cellcolor{red!35} & 0.273 \\

\midrule

&\textbf{Ours} & \textbf{20.32}\cellcolor{red!35} & \textbf{0.596}\cellcolor{red!35} & \textbf{0.350}\cellcolor{red!35} & \textbf{23.72}\cellcolor{red!35} & \textbf{0.723} & \textbf{0.244}\cellcolor{red!35}\\
\bottomrule

\end{tabular}}
\vspace{-2mm}
\caption{\label{table:mip360} Quantitative comparison on  Mip-NeRF360 dataset~\cite{barron2022mipnerf360} for 12 and 24 input views. The best, second-best, and third-best entries are marked in \textcolor{red!70!black}{red}, \textcolor{orange!80!black}{orange}, and \textcolor{yellow!80!black}{yellow} respectively.}

\vspace{-5mm}
\end{table*}

\subsection{Sparse-view Novel View Synthesis}
To combat the ill-posedness of sparse-view reconstruction, a significant body of work has focused on introducing priors through internal regularization. These methods attempt to constrain the solution space using only the information present in the input views or by applying general-purpose heuristics.  Early approaches for NeRF introduced semantic losses from CLIP such as DietNeRF~\cite{Jain_2021_ICCV}, RegNeRF~\cite{Niemeyer2021Regnerf} uses appearance regularization via normalizing flows, while geometric priors like local depth ranking and spatial continuity are used in Sparse-NeRF~\cite{wang2022sparsenerf}. Others focused on training-based strategies, such as FreeNeRF~\cite{Yang2023FreeNeRF} which uses frequency annealing to reduce artifacts, whereas SimpleNeRF~\cite{10.1145/3610548.3618188} and DNGaussian~\cite{li2024dngaussian} utilize depth supervision. More recent methods have adapted these ideas for 3D Gaussian Splatting, using diffusion-based losses for guidance such as Sparse-GS~\cite{xiong2023sparsegs}, or generating pseudo-views to densify the input such as FSGS~\cite{zhu2023FSGS}. CoR-GS~\cite{zhang2024cor} co-regularizes the Gaussians to suppress errors, CoMapGS~\cite{Jang_2025_CVPR} uses uncertainty-aware supervision derived from covisibility maps, Intern-GS~\cite{sun2025interngsvisionmodelguided} uses dense DUSt3R~\cite{dust3r_cvpr24} initialization and foundation-model-guided optimization, while DropGaussian~\cite{park2025dropgaussian} applies dropout to Gaussians to alleviate overfitting. While these approaches improve stability, they are fundamentally data-starved and constrained by regularization. They can refine what is visible but cannot reliably hallucinate the complex, unobserved geometry present in the scene's information voids.

\subsection{External Priors from Auxiliary Vision Models} 

Instead of relying solely on internal regularization, our work pioneers an approach that leverages powerful, pre-trained models from two distinct vision tasks to provide strong external priors.

\paragraph*{Multi-View Stereo (MVS).}
The primary goal of MVS is to estimate dense geometry by establishing robust correspondences across multiple views. Modern MVS architectures, often transformer-based with the advent of Dust3R~\cite{dust3r_cvpr24} and 3R methods~\cite{mast3r_eccv24, must3r_cvpr25, Yang_2025_Fast3R, wang20243d, pow3r_cvpr25}, and line of work~\cite{wang2021deep, wang2023pd, karaev23cotracker, wang2024vggsfm} culminating most recently in VGGT~\cite{wang2025vggt} provide a powerful source of geometric evidence for well-observed regions. Their strength is in producing geometrically accurate outputs where there is sufficient visual information. However, their performance degrades in texture-less areas and fails completely in occluded regions where no correspondences can be found, resulting in incomplete geometric guidance.

\paragraph*{Novel View Synthesis (NVS).}
 In a complementary direction, regressive feedforward NVS Models~\cite{yu2021pixelnerf, charatan23pixelsplat, chen2024mvsplat, ziwen2025llrm, jin2025lvsm} and generative feedforward NVS Models~\cite{liu2023zero1to3,
 zeronvs, 
 wu2023reconfusion, 
 gao2024cat3d, 
 VoletiYBLPTLRJ24, 
 wang2024motionctrl, 
 yu2024viewcrafter, 
watson2025controllingspacetimediffusion} and most recently Stable-Virtual-Camera (SEVA)~\cite{zhou2025stable}   can synthesize photorealistic and semantically coherent images from arbitrary new camera poses. Their strength lies in their ability to plausibly complete a scene, and for diffusion-based generative models, this means hallucinating content for unobserved areas based on learned priors of the visual world. This generative capability, however, comes at the cost of fidelity; they are not constrained by multi-view consistency and can produce plausible-looking "fictions" that contradict the true scene structure.

Thus, these two classes of models are complementary. MVS provides geometrically accurate but incomplete data, while generative NVS provides complete but potentially inaccurate data. Our work introduces a framework to resolve this tension. While previous methods use MVS for direct supervision or generative models for naive inpainting, we repurpose the MVS model as a 3D-aware oracle to validate proposals from the generative model, achieving a synthesis that is both structurally sound and visually complete.

\section{Method}
\label{sec:method}

 \begin{figure*}
\centering

\begin{subfigure}{0.24\textwidth}
    \begin{subfigure}{\textwidth}
        \includegraphics[width=\textwidth]{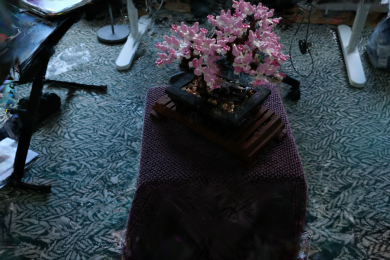}
    \end{subfigure}
    \begin{subfigure}{\textwidth}
        \includegraphics[width=\textwidth]{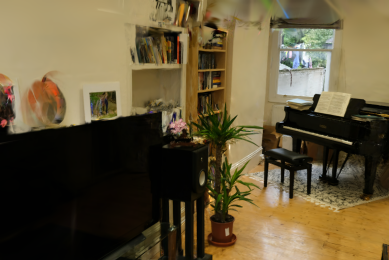}
        
    \end{subfigure}
    \begin{subfigure}{\textwidth}
        \includegraphics[width=\textwidth]{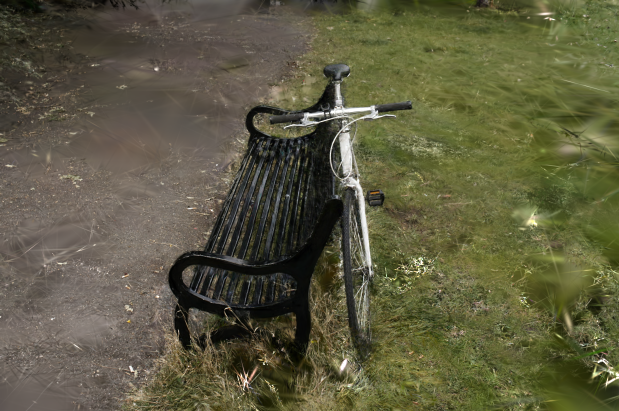} \vspace{-8pt}\\ \scriptsize \centering CoR-GS \cite{zhang2024cor} 
    \end{subfigure}
\end{subfigure}
\begin{subfigure}{0.24\textwidth}
    \begin{subfigure}{\textwidth}
        \includegraphics[width=\textwidth]{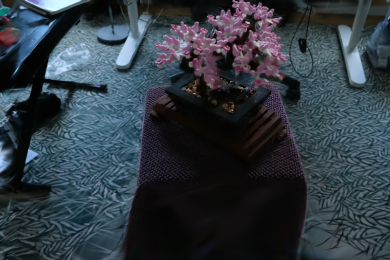}
    \end{subfigure}
    \begin{subfigure}{\textwidth}
        \includegraphics[width=\textwidth]{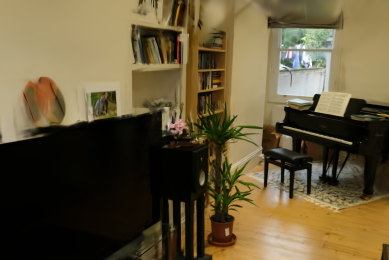}
        
    \end{subfigure}
    \begin{subfigure}{\textwidth}
        \includegraphics[width=\textwidth]{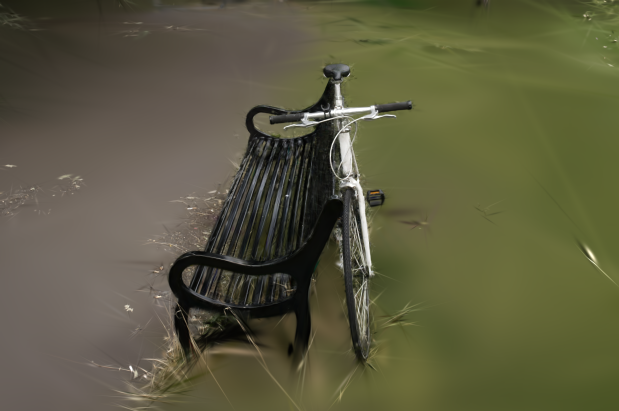} \vspace{-8pt}\\ \scriptsize  \centering DropGaussian \cite{park2025dropgaussian} 
    \end{subfigure}
\end{subfigure}
\begin{subfigure}{0.24\textwidth}
    \begin{subfigure}{\textwidth}
        \includegraphics[width=\textwidth]{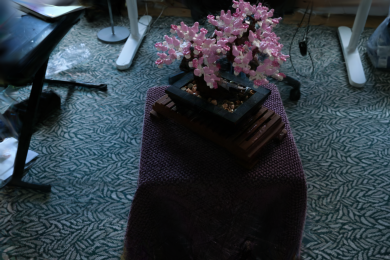}
    \end{subfigure}
    \begin{subfigure}{\textwidth}
        \includegraphics[width=\textwidth]{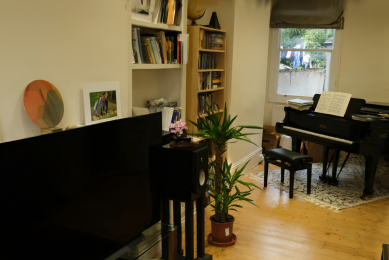}
        
    \end{subfigure}
    \begin{subfigure}{\textwidth}
        \includegraphics[width=\textwidth]{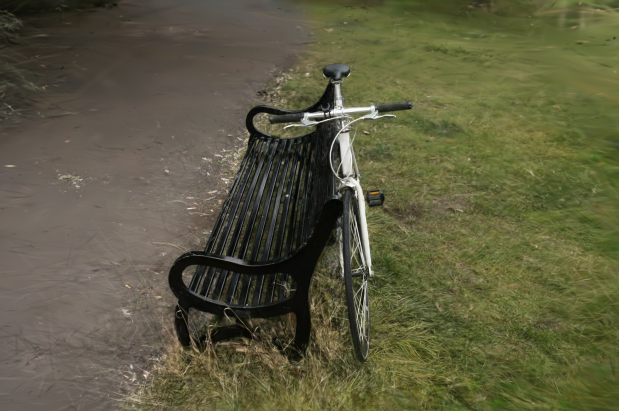} \vspace{-8pt}\\ \scriptsize  \centering OracleGS (Ours)
    \end{subfigure}
\end{subfigure}
\begin{subfigure}{0.24\textwidth}
    \begin{subfigure}{\textwidth}
        \includegraphics[width=\textwidth]{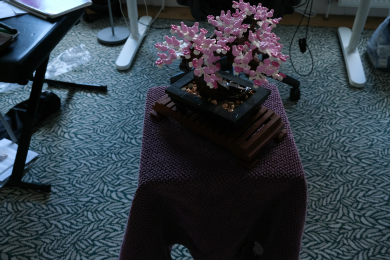}
    \end{subfigure}
    \begin{subfigure}{\textwidth}
        \includegraphics[width=\textwidth]{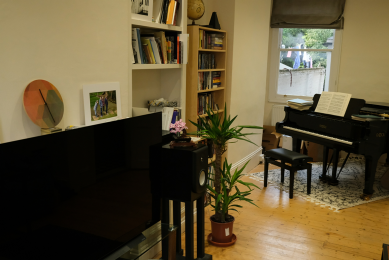}
        
    \end{subfigure}
    \begin{subfigure}{\textwidth}
        \includegraphics[width=\textwidth]{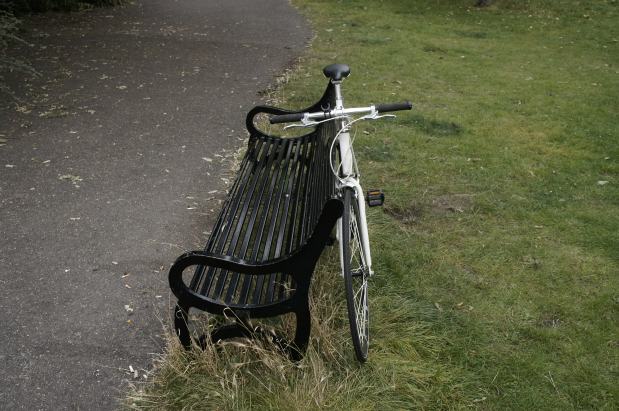} \vspace{-8pt}\\ \scriptsize \centering Ground Truth
    \end{subfigure}
\end{subfigure}

\caption{\textbf{Visual comparison with state-of-the-art methods on the Mip-NeRF360~\cite{barron2022mipnerf360} dataset}. Our method, OracleGS, demonstrates superior handling of common failure modes. \textbf{Top row (Bonsai, 12 views):} OracleGS accurately reconstructs the challenging carpet texture and background regions while competing methods produce noticeable artifacts. \textbf{Middle row (Room, 12 views):} Our method avoids the distortions present in other reconstructions. \textbf{Bottom row (Bicycle, 24 views):} Our approach strikes a balance between detail and smoothness, preventing the noisy overfitting seen in CoR-GS~\cite{zhang2024cor} and the oversmoothing that erases fine details in DropGaussian~\cite{park2025dropgaussian}.}
\label{fig:mipnerf360_results} 
\end{figure*}

\label{sec:preliminaries}
\subsection{Preliminaries}
3D Gaussian Splatting (3DGS)~\cite{kerbl3Dgaussians} is an explicit scene representation that uses a collection of 3D Gaussians for high-fidelity, real-time novel view synthesis. Each Gaussian is defined by a set of optimizable attributes: a position (mean) $\boldsymbol{\mu} \in \mathbb{R}^3$, a covariance matrix $\boldsymbol{\Sigma} \in \mathbb{R}^{3\times3}$, an opacity $\alpha \in [0, 1]$, and a color $\mathbf{c} \in \mathbb{R}^3$. To model view-dependent effects, the color is typically represented by Spherical Harmonics (SH) coefficients. For efficient optimization, the covariance matrix $\boldsymbol{\Sigma}$ is parameterized by a 3D scaling vector $\mathbf{s}$ and a rotation quaternion $\mathbf{q}$. To render a novel view, the 3D Gaussians are projected onto the 2D image plane and then blended in depth-sorted order. The final color $\mathbf{C}$ for a pixel is computed via alpha-blending:
\begin{equation}
    \mathbf{C} = \sum_{i=1}^{N} T_i \alpha_i \mathbf{c}_i,
    \label{eq:blending}
\end{equation}
where $N$ is the number of Gaussians overlapping the pixel, sorted by depth. The color of the $i$-th Gaussian is $\mathbf{c}_i$, and its opacity is $\alpha_i$. The accumulated transmittance $T_i$ ensures that Gaussians closer to the camera have a greater contribution.

While 3DGS achieves state-of-the-art results, its performance is contingent upon a dense set of input views that provide robust geometric and photometric constraints. In the sparse-view setting, the optimization becomes severely under-constrained, often leading to undesirable artifacts such as floater artifacts and incomplete, blurry renderings. 

We introduce \textbf{OracleGS}, with a ``propose-and-validate'' framework that synergizes generative and regressive frameworks. First, a generative model \emph{proposes} a complete scene by synthesizing novel views. An MVS model, repurposed as a 3D-aware oracle, then \emph{validates} these proposals by quantifying their 3D uncertainty. This signal guides the optimization of a 3DGS~\cite{kerbl3Dgaussians} representation, optimized using 3DGS-MCMC~\cite{kheradmand20243d}, which is initialized from a COLMAP~\cite{schoenberger2016sfm} point cloud generated using ground-truth poses. Our method is illustrated in Figure~\ref{fig:pipeline}.

\subsection{View Augmentation \& Uncertainty Estimation}
\label{sec:propose_validate}

The foundation of our method is a new architecture that assigns synergistic roles to its generative and regressive components via the ``Propose-and-validate'' framework.

\paragraph{View Augmentation via Generative NVS.} The first stage addresses the information gap inherent in sparse inputs. We use a pre-trained, 3D-aware diffusion model (SEVA)~\cite{zhou2025stable} to \emph{propose} a complete version of the scene. Given the sparse set of $N$ ground-truth (GT) images $\{I_i\}_{i=1}^N$ and their corresponding camera poses $\{P_i\}_{i=1}^N$, the model generates a dense set of $M$ synthetic images $\{I'_j\}_{j=1}^M$ from new query camera poses $\{P'_j\}_{j=1}^M$. Though the generative model creates novel views, they are not geometrically perfect. The purpose of this stage is to provide a coarse proposal for the scene's appearance and structure, effectively filling unobserved regions. However, these synthetic views $\{I'_j\}$ are a rich but potentially inconsistent source of data; directly using them without refinement would propagate geometric inaccuracies and visual artifacts, leading to a suboptimal final reconstruction. The subsequent stages are designed to mitigate these issues.

\paragraph{Uncertainty Estimation via a Repurposed MVS Oracle.}
Proposed synthetic views contain inconsistencies, and using them naively destabilizes the training process. To validate the structural integrity of the proposed synthetic views, we introduce a \textit{3D-Aware Oracle} whose task is to assess their 3D uncertainty. While traditional methods leverage MVS models for direct supervision via their final depth or point cloud outputs, these are often noisy and incomplete in sparse-view settings, providing unreliable guidance. We circumvent this limitation by extracting a more reliable signal directly from the MVS model's intrinsics. Inspired by recent findings that attention maps in vision transformers encode rich structural cues~\cite{chen2025easi3r}, we repurpose the \textbf{global-attention maps} from a transformer-based MVS model, VGGT~\cite{wang2025vggt} as a powerful proxy for 3D uncertainty. Specifically, a high global-attention score between views signifies strong multi-view consistency and thus high 3D confidence. Conversely, a low score suggests high 3D uncertainty, effectively identifying generative hallucinations that lack robust multi-view support. This fine-grained uncertainty signal is then used to guide the final optimization.
\vspace{-4mm}
\paragraph{3D-Aware Oracle Implementation.}
To implement the geometric oracle $\Phi$, we first partition the set of synthetic views $\{I'_j\}_{j=1}^M$ into disjoint chunks $\{C_k\}$. This is necessary because the GPU memory requirements of the MVS model~\cite{wang2025vggt} scales with the number of input views. Each chunk $C_k$ is independently processed by the MVS model to assess the 3D uncertainty of the generated scene proposal. Let $\Psi^{(l)}(I'_j | C_k)$ be an operator that extracts the global attention maps from view $I'_j \in C_k$ from layer $l$ of the MVS model, computed with respect to the other views within its own chunk $C_k$. The final uncertainty map $U_j \in [0, 1]^{H \times W}$ is a weighted average of these layer normalized attention maps from a set of layers $\mathcal{L}$:
\begin{equation}
    U_j = \sum_{l \in \mathcal{L}} w_l \cdot \Psi^{(l)}(I'_j | C_k), \quad \text{where } I'_j \in C_k.
\end{equation}
In our implementation, we use layers $\mathcal{L} = \{0, 22\}$, reference view $I_j = I_0$, with corresponding weights $w_l= \{\frac{1}{4},\frac{3}{4}\}$. We provide an analysis for this layer selection in the supplementary material. This map serves as the oracle's final judgment, with $U_j(u,v) \approx 1$ signifying high confidence, low uncertainty and $U_j(u,v) \approx 0$ signifying low confidence, high uncertainty.

\subsection{Uncertainty-Enforced 3DGS Optimization}
Having generated the synthetic proposals and the corresponding per-pixel uncertainty maps, we now detail how these components are integrated to supervise the 3DGS optimization. The final stage uses sparse GT training views, synthetic proposals from non-test viewpoints, and the oracle's uncertainty maps to optimize a 3DGS scene.

\vspace{-3mm}
\paragraph{Ground-Truth and Depth Loss ($\mathcal{L}_{\text{GT}}$).} For any sampled GT view $I_i$, and predicted view $\hat{I}_i$ the loss combines a standard photometric term with a depth term.
\begin{equation}
    \mathcal{L}_{\text{GT}} = \mathcal{L}_{\text{photo}} + \lambda_{\text{depth}}(t) \mathcal{L}_{\text{depth}}
    \label{eq:gt_loss}
\end{equation}
The photometric loss, $\mathcal{L}_{\text{photo}}$, is a combination of L1 and SSIM:
\begin{equation}
    \mathcal{L}_{\text{photo}} = (1 - \lambda_{\text{ssim}}) \mathcal{L}_1(\hat{I}_i, I_i) + \lambda_{\text{ssim}} \mathcal{L}_{\text{SSIM}}(\hat{I}_i, I_i)
    \label{eq:photo_loss}
\end{equation}
The depth loss for 3DGS $\mathcal{L}_{\text{depth}}$, as established in Hierarchical 3DGS~\cite{hierarchicalgaussians24},  is applied when reliable monocular depth priors are available for a GT view. It is a masked L1 loss between the rendered inverse depth $\hat{D}_i$ and the provided inverse depth map $D_i$:
\begin{equation}
    \mathcal{L}_{\text{depth}} = \| (\hat{D}_i - D_i) \odot M_i \|_1
    \label{eq:depth_loss}
\end{equation}
where $M_i$ is a mask indicating reliable depth regions and $\odot$ indicates element-wise multiplication. The weight $\lambda_{\text{depth}}(t)$ is annealed with an exponential schedule with initial and final weights $\lambda_{depth_{0}}=1$, $\lambda_{depth_{1}}=0.01$ respectively.

\paragraph{Uncertainty-Weighted Synthetic Loss ($\mathcal{L}_{\text{synth}}$).} For a sampled synthetic view $I'_j$, and predicted synthetic novel view $\hat{I}'_j$, the loss is modulated by the oracle's uncertainty map $U_j$. Standard photometric losses are highly sensitive to pixel-level artifacts in generated images. We therefore incorporate a perceptual LPIPS~\cite{zhang2018lpips} loss, which is more robust to minor texture variations. The total synthetic loss is:
\begin{equation}
\begin{split}
    \mathcal{L}_{\text{synth}} = & (1 - \lambda'_{\text{SSIM}} - \lambda_{\text{LPIPS}}) \cdot \| (\hat{I}'_j - I'_j) \odot U_j \|_1 \\
    & + \lambda'_{\text{SSIM}} \cdot \| (1 - \text{SSIM}(\hat{I}'_j, I'_j)) \odot U_j \|_1 \\
    & + \lambda_{\text{LPIPS}} \cdot \bar{U}_j \cdot \mathcal{L}_{\text{LPIPS}}(\hat{I}'_j, I'_j)
\end{split}
\label{eq:synth_loss}
\end{equation}
Here, the L1 loss is weighted on a per-pixel basis by the uncertainty map $U_j$. We perform an element-wise multiplication of the map SSIM ~\cite{wang2004ssim} creates within each patch with our uncertainty map $U_j$ before averaging to a final loss value. This mechanism forces the optimization to prioritize structural consistency in regions deemed reliable by the oracle, while effectively down-weighting the contribution from patches in uncertain or artifact-prone areas. The LPIPS~\cite{zhang2018lpips} loss, which operates on image patches, is modulated by a single scalar value $\bar{U}_j = \text{mean}(U_j)$, representing the overall geometric integrity of the entire synthetic view. This prevents low-uncertainty patches from subsidizing high-uncertainty ones in the perceptual loss, ensuring that low uncertainty images contribute significantly to the final appearance. We inherit the regularization terms on opacity and scale ($\mathcal{L}_{\text{reg}}$) from 3DGS-MCMC~\cite{kheradmand20243d} to encourage a compact representation. The total loss becomes:
\begin{equation}
    \mathcal{L} = \mathcal{L}_{\text{GT}} + \mathcal{L}_{\text{synth}} + \mathcal{L}_{\text{reg}}
    \label{eq:total_loss}
\end{equation}

\subsection{Progressive Augmentation Schedule}

To further stabilize training, we employ a dynamic training curriculum, governed by a scaled and shifted Beta distribution, to schedule the sampling probability of synthetic views over time. This schedule is designed to use the synthetic data as an intermediary scaffold for reconstruction. Initially, the probability of sampling synthetic views is low, allowing the initial point cloud to guide the scene. Afterwards, the probability of selecting a synthetic view increases, allowing their dense geometric and appearance information to rapidly structure the scene, including the unobserved regions, providing a coherent foundation that the sparse initial point cloud cannot. As optimization progresses and the learning rate anneals, the sampling probability of a synthetic view decays, reducing the influence of the potentially imperfect synthetic data, allowing the model to dedicate its capacity to fitting the fine-grained, high-fidelity details present primarily in the ground-truth images. This curriculum ensures that the generative prior provides its main contribution of global structure when most needed, before gracefully yielding to the ground-truth data for final refinement. Please see the experimental settings and supplementary material for details.

\section{Experiments}

\begin{figure}[t!]
\centering
\setlength{\tabcolsep}{1pt} 

\begin{subfigure}{0.24\columnwidth}
    \includegraphics[width=\textwidth]{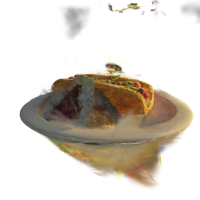}
\end{subfigure}\hfill
\begin{subfigure}{0.24\columnwidth}
    \includegraphics[width=\textwidth]{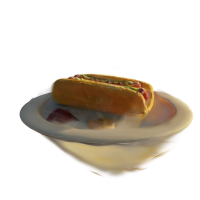}
\end{subfigure}\hfill
\begin{subfigure}{0.24\columnwidth}
    \includegraphics[width=\textwidth]{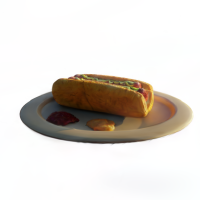}
\end{subfigure}\hfill
\begin{subfigure}{0.24\columnwidth}
    \includegraphics[width=\textwidth]{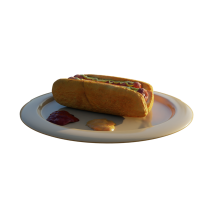}
\end{subfigure}

\begin{subfigure}{0.24\columnwidth}
    \includegraphics[width=\textwidth]{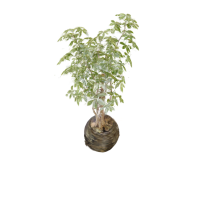}
\end{subfigure}\hfill
\begin{subfigure}{0.24\columnwidth}
    \includegraphics[width=\textwidth]{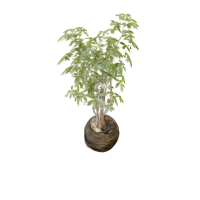}
\end{subfigure}\hfill
\begin{subfigure}{0.24\columnwidth}
    \includegraphics[width=\textwidth]{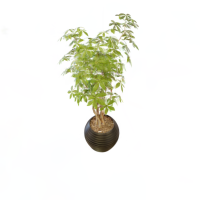}
\end{subfigure}\hfill
\begin{subfigure}{0.24\columnwidth}
    \includegraphics[width=\textwidth]{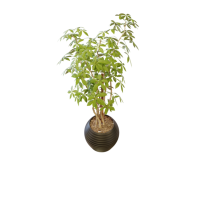}
\end{subfigure}

\begin{subfigure}{0.24\columnwidth}
    \includegraphics[width=\textwidth]{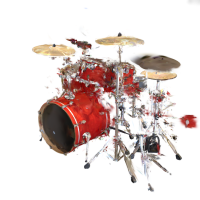}
\end{subfigure}\hfill
\begin{subfigure}{0.24\columnwidth}
    \includegraphics[width=\textwidth]{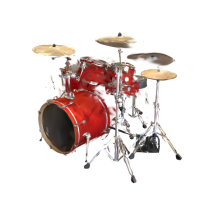}
\end{subfigure}\hfill
\begin{subfigure}{0.24\columnwidth}
    \includegraphics[width=\textwidth]{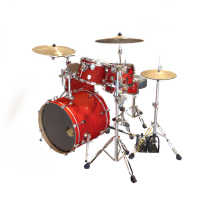}
\end{subfigure}\hfill
\begin{subfigure}{0.24\columnwidth}
    \includegraphics[width=\textwidth]{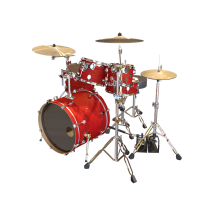}
\end{subfigure}

\subcaptionbox*{\parbox[t]{0.24\columnwidth}{\centering \scriptsize CoR-GS~\cite{zhang2024cor}}}{\hspace{0.24\columnwidth}}%
\hfill
\subcaptionbox*{\parbox[t]{0.24\columnwidth}{\centering \scriptsize DropGaussian~\cite{park2025dropgaussian}}}{\hspace{0.24\columnwidth}}%
\hfill
\subcaptionbox*{\parbox[t]{0.24\columnwidth}{\centering \scriptsize \textbf{Ours}}}{\hspace{0.24\columnwidth}}%
\hfill
\subcaptionbox*{\parbox[t]{0.24\columnwidth}{\centering \scriptsize Ground Truth}}{\hspace{0.24\columnwidth}}

\caption{\textbf{Visual comparison with state-of-the-art methods on the NeRF Synthetic~\cite{mildenhall2020nerf} dataset}. Our method consistently produces higher-fidelity reconstructions across diverse and challenging scenes compared to prior work. \textbf{Top row (Hotdog):} OracleGS eliminates the "floater" artifacts  present in competing methods, preserving details on the condiments and plate. \textbf{Middle row (Ficus):} Our method reconstructs the intricate vase structures where competing methods fail. \textbf{Bottom row (Drums):} Our method reconstructs the thin structures of the drum kit's stands and cymbals, which are fragmented or missing in other reconstructions.}
\label{fig:blender_results} 
\end{figure}

\subsection{Experimental Settings}

\paragraph*{Datasets} We conduct our experiments on two datasets,
Mip-NeRF360 ~\cite{barron2022mipnerf360} dataset, which features seven challenging 360$^\circ$ scenes;
and NeRF Synthetic~\cite{mildenhall2020nerf} dataset which features eight scenes of 360$^\circ$ pathtraced objects with realistic non-Lambertian materials.
For the Mip-NeRF 360 dataset, we use every 8th image as testing view and evenly sample 12 or 24 views from the remaining views as the training set, and all input images are downsampled to $\tfrac{1}{4}$ of the original height and width. 
For the NeRF Synthetic dataset, we follow the protocols established in DietNeRF~\cite{Jain_2021_ICCV} and FreeNeRF~\cite{Yang2023FreeNeRF} by using 8 images for training and 25 for testing, all input images are downscaled to $\tfrac{1}{2}$ of the original height and width.
\vspace{-2mm}
\paragraph*{Implementation Details}
We inherit the standard parameters from~\cite{kheradmand20243d} and additionally set 
\(\lambda'_{\text{SSIM}} = \lambda_{\text{LPIPS}} = 0.3\) and train up to 22,000 iterations. During training, the probability of sampling a synthetic image is modulated over time according to 
\(p_{\text{synthetic}} \sim 0.1 + 20\,\mathrm{Beta}(t; \alpha=2,\beta=4)\),  where \(t\) is the normalized training progress between 0 and 1. 
We use Depth-Anything V2~\cite{depth_anything_v2} to predict pseudo inverse depth and generate synthetic images 
\({I'}\) using \textsc{Seva}~\cite{zhou2025stable}.
We report PSNR, SSIM~\cite{wang2004ssim}, and LPIPS~\cite{zhang2018lpips} to evaluate reconstruction performance quantitatively. We used a single A40 GPU for all experiments. For more implementation details, please refer to the supplementary material.

\subsection{Comparison}
\begin{table}[t!]
\centering
\adjustbox{max width=\columnwidth}{%
\centering
\fontsize{9.5pt}{11pt}\selectfont
\renewcommand{\arraystretch}{1.1} %
\begin{tabular}{c l| c c c}
\toprule
\multicolumn{2}{c}{Methods} & PSNR($\uparrow$) & SSIM($\uparrow$) & LPIPS ($\downarrow$) \\ 
\midrule

& RegNeRF~\cite{Niemeyer2021Regnerf} & 23.86 &  0.852 & 0.105 \\
& FreeNeRF~\cite{Yang2023FreeNeRF} & 24.26 &  0.883 & 0.098 \\
& SparseNeRF~\cite{wang2022sparsenerf} & 24.04 &  0.876 & 0.113 \\ \midrule
& 3DGS~\cite{kerbl3Dgaussians} & 21.56 & 0.847 & 0.130 \\
& DNGaussian~\cite{li2024dngaussian} & 24.31 & 0.886 & 0.088 \\
& FSGS~\cite{zhu2023FSGS} & 24.64\cellcolor{yellow!35} &  0.895\cellcolor{yellow!35} & 0.095 \\
& CoR-GS~\cite{zhang2024cor} & 24.43 & 0.896\cellcolor{orange!35} & 0.084\cellcolor{orange!35} \\
& DropGaussian~\cite{park2025dropgaussian} & 25.42\cellcolor{red!35} & 0.888 & 0.089\cellcolor{yellow!35} \\ 
\midrule

& \textbf{Ours} & \textbf{24.75}\cellcolor{orange!35} & \textbf{0.905}\cellcolor{red!35} & \textbf{0.067}\cellcolor{red!35} \\

\bottomrule
\end{tabular}}

\caption{\label{table:nerf_synth} \textbf{Quantitative comparison on the Blender~\cite{mildenhall2020nerf}} dataset for 8 input views. Best, second, and third results are marked in \textcolor{red!70!black}{red}, \textcolor{orange!80!black}{orange}, and \textcolor{yellow!80!black}{yellow} respectively.}

\end{table}
\paragraph*{Evaluation on Mip-NeRF 360}
We conduct our evaluation on the challenging Mip-NeRF 360 dataset~\cite{barron2022mipnerf360}, which features 7 complex, unbounded indoor and outdoor 360\textdegree~scenes with significant occlusions, making it an ideal benchmark for sparse-view reconstruction. We evaluate all scenes for 12-view setting at 15k and all scenes for 24-view at 22k iterations. As shown in Table~\ref{table:mip360}, our method, \textbf{OracleGS}, establishes a new state-of-the-art in the extremely sparse \textbf{12-view} setting, outperforming all prior work across all three metrics. This strong performance is particularly notable given the framework's robustness to generative failures. 
In the \textbf{24-view} setting, the reliance on generative priors diminishes, but \textbf{OracleGS} remains highly competitive, achieving the best performance for PSNR and LPIPS by a margin. This demonstrates that while our approach provides the most significant advantage in the critically underdetermined low-data regime, it maintains the state-of-the-art performance as view density increases. For a fair comparison, we report results for DropGaussian~\cite{park2025dropgaussian} using the standard $\tfrac{1}{4}$ image downsampling, consistent with prior work, as the official paper evaluates using a non-standard $\tfrac{1}{8}$ resolution. Figure~\ref{fig:mipnerf360_results} shows our qualitative results on Mip-NeRF 360~\cite{barron2022mipnerf360} dataset. Please refer to the supplementary material for per scene results on Mip-NeRF 360. 
\vspace{-4mm}
\paragraph{Evaluation on NeRF Synthetic}
We conduct our secondary evaluation on NeRF Synthetic dataset~\cite{mildenhall2020nerf}.
Quantitative results on the NeRF Synthetic dataset with 8 training views are reported in Table~\ref{table:nerf_synth}. Our method achieves the best scores in SSIM and LPIPS with second best results in PSNR. This indicates that our method preserves finer, perceptually important details at the expense of minute pixel-level deviations that PSNR is sensitive to as demonstrated in Figure~\ref{fig:blender_results} in our qualitative results on NeRF Synthetic~\cite{mildenhall2020nerf} dataset.

\begin{figure}[t]
    \centering
    \setlength{\tabcolsep}{0.7pt} %
    \renewcommand{\arraystretch}{0.1} %
    \begin{tabular}{@{}c@{\hspace{6pt}}cc|ccc@{}}
        \rotatebox{90}{\scriptsize \hspace{1em}Synth.} &
        \includegraphics[width=0.14\linewidth]{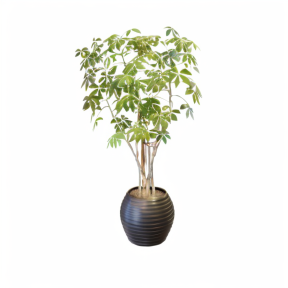} &
        \includegraphics[width=0.14\linewidth]{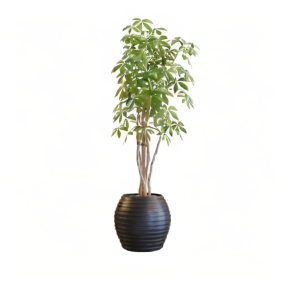} &
        \includegraphics[width=0.215\linewidth]{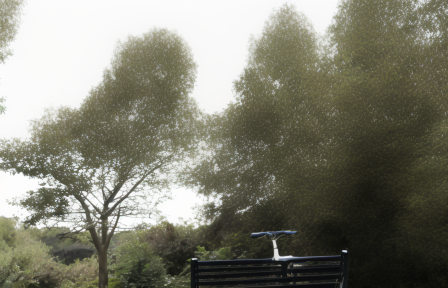}&
        \includegraphics[width=0.215\linewidth]{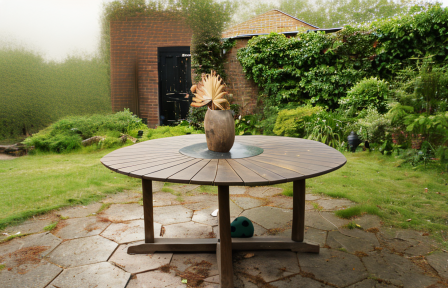} &
        \includegraphics[width=0.215\linewidth]{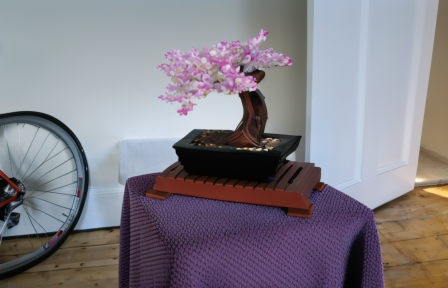} \\[0.2ex]  %

        \rotatebox{90}{\scriptsize\hspace{0.1em} Attn. Map} &
        \includegraphics[width=0.14\linewidth]{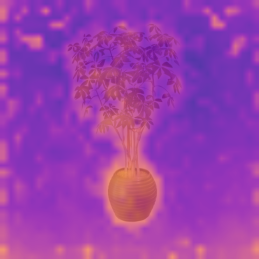} &
        \includegraphics[width=0.14\linewidth]{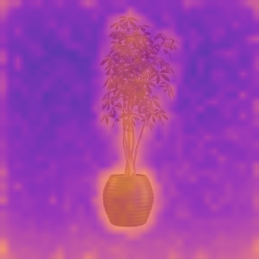} &
        \includegraphics[width=0.215\linewidth]{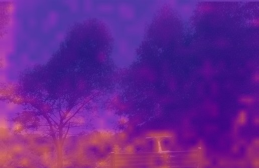} &
        \includegraphics[width=0.215\linewidth]{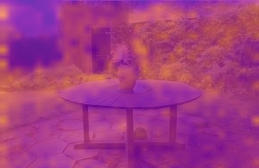} &
        \includegraphics[width=0.215\linewidth]{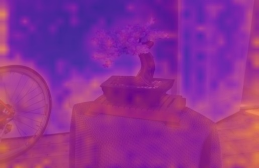} \\[0.2ex]

        \rotatebox{90}{\scriptsize\hspace{0.7em} GT} &
        \includegraphics[width=0.14\linewidth]{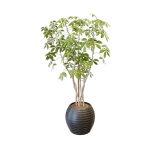} &
        \includegraphics[width=0.14\linewidth]{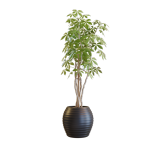} &
        \includegraphics[width=0.215\linewidth]{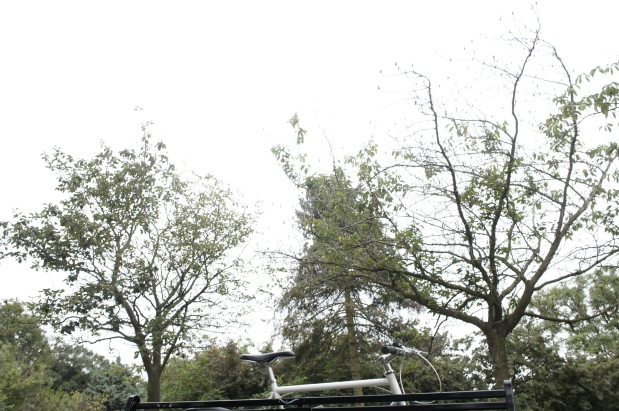} &
        \includegraphics[width=0.215\linewidth]{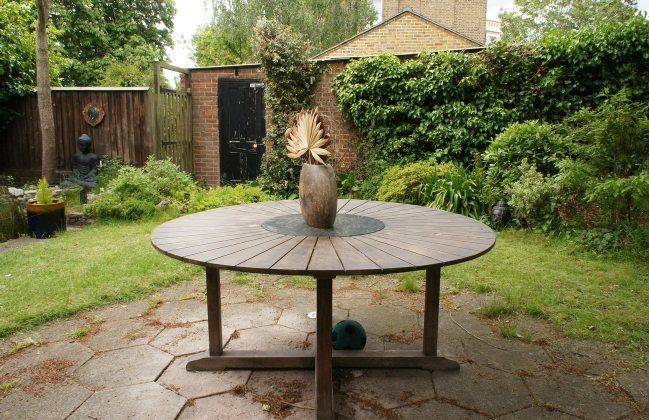} &
        \includegraphics[width=0.215\linewidth]{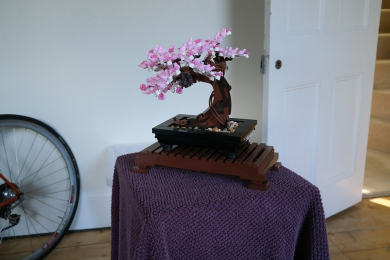} \\[0.2ex]
        
        & \multicolumn{2}{c|}{\parbox[c]{0.28\linewidth}{\centering \scriptsize Object Level 3D Inconsistencies}} &
        \parbox[c]{0.215\linewidth}{\centering \scriptsize Underobserved Regions} &
        \parbox[c]{0.215\linewidth}{\centering \scriptsize Background Inconsistencies} &
        \parbox[c]{0.215\linewidth}{\centering \scriptsize Texture-less regions} \\
    \end{tabular}
    \caption{\textbf{Our 3D-aware Oracle quantifies diverse sources of 3D uncertainty.} We visualize extracted uncertainty maps (middle row) on synthetic images from the generative model using global-attention layers from the repurposed MVS model~\cite{wang2025vggt} \emph{normalized} from \BGcolor{fff77f}{o}%
\BGcolor{ffdf5e}{n}%
\BGcolor{ffb94c}{e}%
\BGcolor{ff933a}{ }%
\BGcolor{ff6d28}{t}%
\BGcolor{f54046}{o}%
\BGcolor{c93c8e}{ }%
\BGcolor{9c3ca9}{z}%
\BGcolor{7b44b2}{e}%
\BGcolor{5d4ec2}{r}%
\BGcolor{4657c7}{o}%
where low uncertainty is shown in yellow and high uncertainty is shown in purple. Each column demonstrates the oracle's ability to identify a specific failure mode in the synthetic proposals by comparing against the ground truth.}
\label{fig:oracle_uncertainty_quantification}
\end{figure}

\subsection{Analysis of the 3D-Aware Oracle}
To provide a deeper insight into the effectiveness of our proposed 3D-aware oracle, we visualize its output on several challenging scenes. As shown in Figure~\ref{fig:oracle_uncertainty_quantification}, the oracle is remarkably effective in identifying diverse sources of 3D uncertainty. For instance, the \textbf{Ficus} scene (left) demonstrates the oracle's discriminative power: it assigns low uncertainty to a 3D \emph{consistent} proposal on the right while correctly identifying a hallucinated, \emph{inconsistent} leaf structure in another proposal on the left for the same scene. In the \textbf{Bicycle} scene (middle left), the oracle identifies regions of high \emph{epistemic uncertainty} that are simply \emph{under-observed} in the sparse input views. By assigning high uncertainty to the background, our method is prevents overfitting to a potentially inaccurate generative completion. The oracle is also effective at identifying inconsistencies at various distances, as shown in the \textbf{Garden} scene (middle right), where it successfully flags inconsistencies in the distant background foliage on the top left part of the image. Finally, the \textbf{Bonsai} scene (right) shows the oracle correctly identifying \emph{textureless} surfaces as inherently uncertain. This is advantageous during optimization because it prevents the model from wasting capacity trying to perfectly replicate potentially noisy generative details on these low-information surfaces, allowing it to focus on reconstructing the high-confidence foreground object. This
 uncertainty signal is what allows our uncertainty-weighted loss to ground the generative prior.

\subsection{Ablation Study}
\label{sec:ablation_study}

\begin{table}[t!]
  \centering
  \resizebox{\linewidth}{!}{%
    \footnotesize 
    \setlength{\tabcolsep}{3.5pt} 
    \renewcommand{\arraystretch}{1.1} %
    
    \begin{tabular}{l | S[table-format=2.3] S[table-format=1.3] S[table-format=1.3] | S[table-format=2.3] S[table-format=1.3] S[table-format=1.3]}
      \toprule
      \multirow{2}{*}[-0.5ex]{\textbf{Method}} & \multicolumn{3}{c|}{\textbf{12-view}} & \multicolumn{3}{c}{\textbf{24-view}} \\
      \cmidrule(lr){2-4} \cmidrule(lr){5-7}
      & {PSNR $\uparrow$} & {SSIM $\uparrow$} & {LPIPS $\downarrow$} & {PSNR $\uparrow$} & {SSIM $\uparrow$} & {LPIPS $\downarrow$} \\
      \midrule

      Baseline~\cite{kheradmand20243d} & 14.10 & 0.285 & 0.560 & 17.95 & 0.479 & 0.409 \\
      + Syn. Views & 16.72 & 0.382 & 0.596 & 16.88 & 0.428 & 0.565 \\
      + Schedule & 17.32 & 0.433 & 0.515 & 29.85 & 0.585 & 0.362 \\
      + LPIPS Loss & 17.60 & 0.446 & 0.459 & 21.34 & 0.598 & 0.328 \\
      + SfM Init. & 19.96 & 0.583 & 0.362 & 22.88 & 0.691 & 0.284 \\
      + Depth Loss & 20.28 & 0.592 & 0.354 & 23.64 & 0.717 & 0.258 \\
      \midrule
      \textbf{+ Unc. Guide} & \textbf{20.32} & \textbf{0.596} & \textbf{0.350} & \textbf{23.72} & \textbf{0.723} & \textbf{0.244} \\
      \bottomrule
    \end{tabular}%
  } %
  \caption{\textbf{Ablation study on Mip-NeRF 360~\cite{barron2022mipnerf360}}, showing the progressive impact of each component for 12-view and 24-view settings. Our full method is shown in bold.}
  \label{tab:ablation_full}
\end{table}

We conduct a comprehensive cumulative ablation study on the Mip-NeRF 360 benchmark~\cite{barron2022mipnerf360}. 
We first demonstrate that naively incorporating \textbf{Synthetic Views (+Syn. Views)}  into the training doesn't improve the performance significantly in the sparse-view case as demonstrated by the 12-view ablations. As the number of training images increase, naive incorporation of synthetic images actually hurts performance. This confirms that raw generative artifacts can corrupt the optimization. To address this instability, we introduce a \textbf{Progressive Augmentation Schedule (+Schedule)}. This schedule provides a significant performance recovery, and it's critical role is visually demonstrated in the supplementary material, where omitting it leads to catastrophic geometric collapse. We further stabilize training with \textbf{LPIPS Loss}, providing tolerance to minor generative artifacts that pixel-wise losses penalize.
We then incorporate established techniques like \textbf{SfM Initialization} (\textbf{+SfM Init.}) and \textbf{Depth Regularization} (\textbf{+Depth Loss}), which, as expected, provide improvements by grounding the optimization in a plausible geometric state. Finally, we introduce \textbf{uncertainty-guided loss} (\textbf{+Unc. Guide}). Despite being applied after several strong stabilization components which have already significantly improved the baseline, it improves the quantitative and qualitative results. The uncertainty guidance acts as an intelligent filter, selectively integrating the now-stabilized synthetic data based on its 3D uncertainty to achieve our final state-of-the-art performance.

\begin{figure}[t!]
  \centering
  \setlength{\tabcolsep}{2pt} 
  \begin{subfigure}[b]{0.32\columnwidth}
    \includegraphics[width=\textwidth]{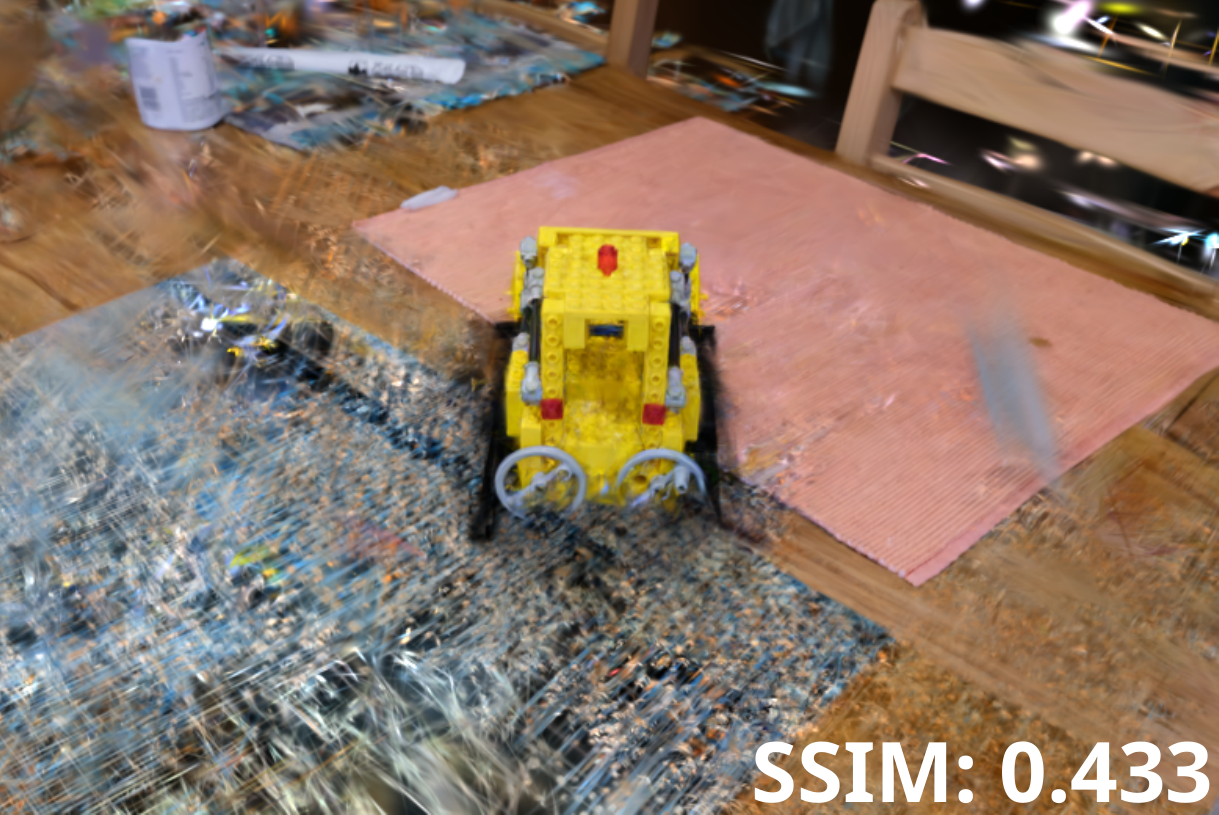}
    \caption*{\parbox[t]{\linewidth}{\centering\scriptsize Baseline~\cite{kheradmand20243d}}}
    \label{fig:ablation_a}
  \end{subfigure}
  \hfill 
  \begin{subfigure}[b]{0.32\columnwidth}
    \includegraphics[width=\textwidth]{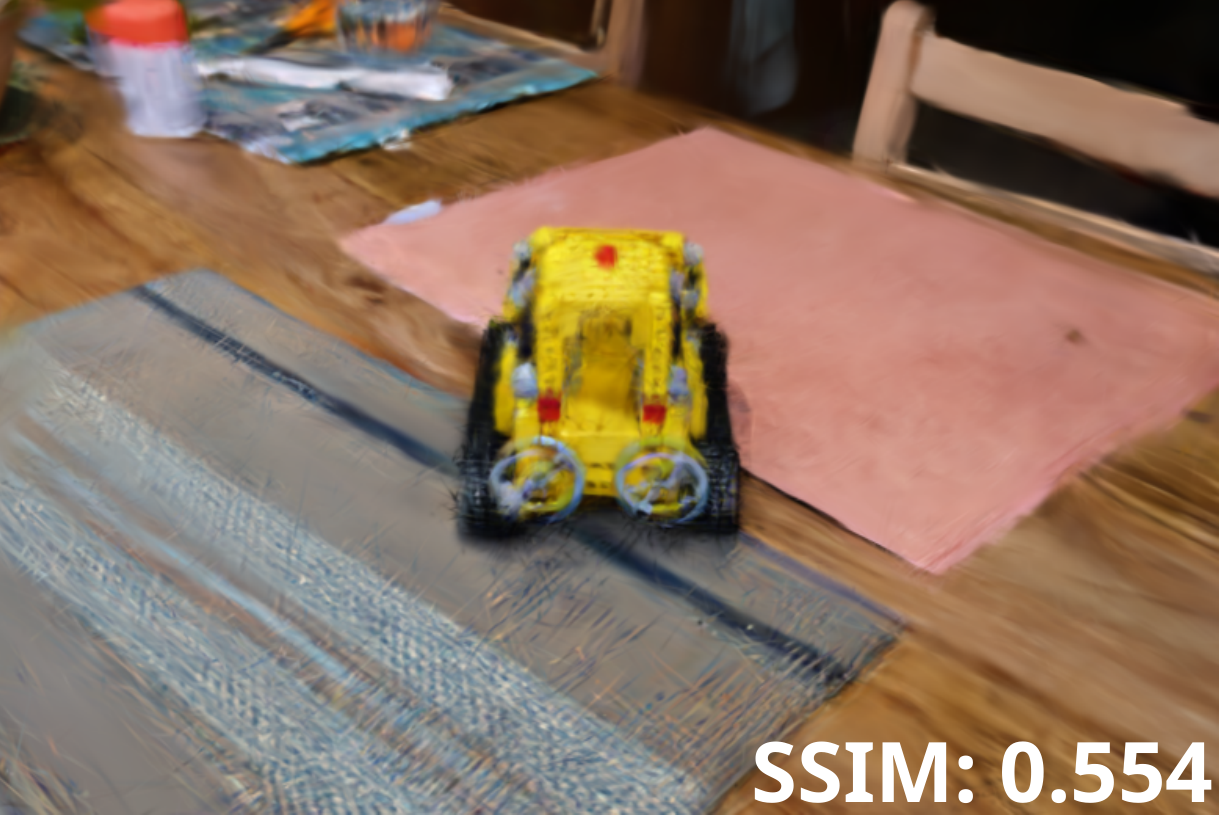}
    \caption*{\parbox[t]{\linewidth}{\centering\scriptsize + Synthetic Views}}
    \label{fig:ablation_b}
  \end{subfigure}
  \hfill
  \begin{subfigure}[b]{0.32\columnwidth}
    \includegraphics[width=\textwidth]{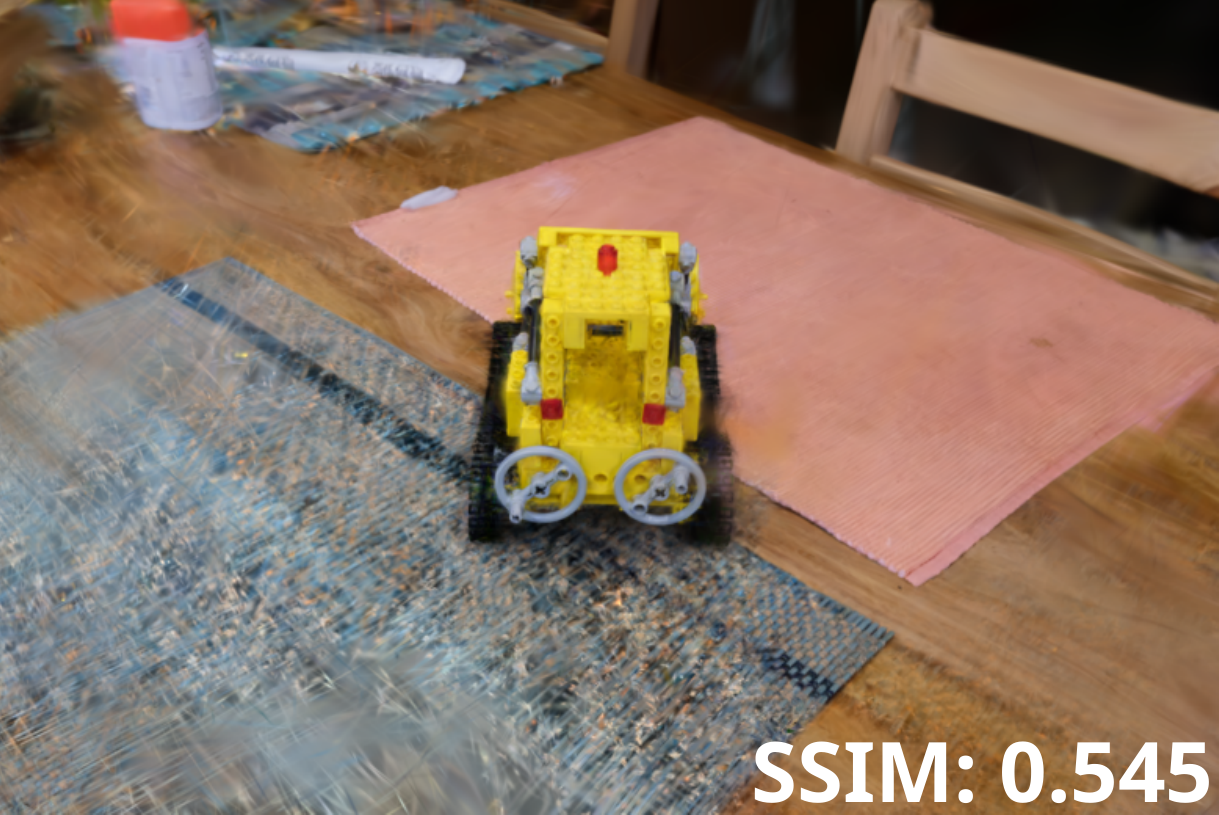}
    \caption*{\parbox[t]{\linewidth}{\centering\scriptsize + Schedule }}
    \label{fig:ablation_c}
  \end{subfigure}

  \vspace{1mm} 
  \begin{subfigure}[b]{0.32\columnwidth}
    \includegraphics[width=\textwidth]{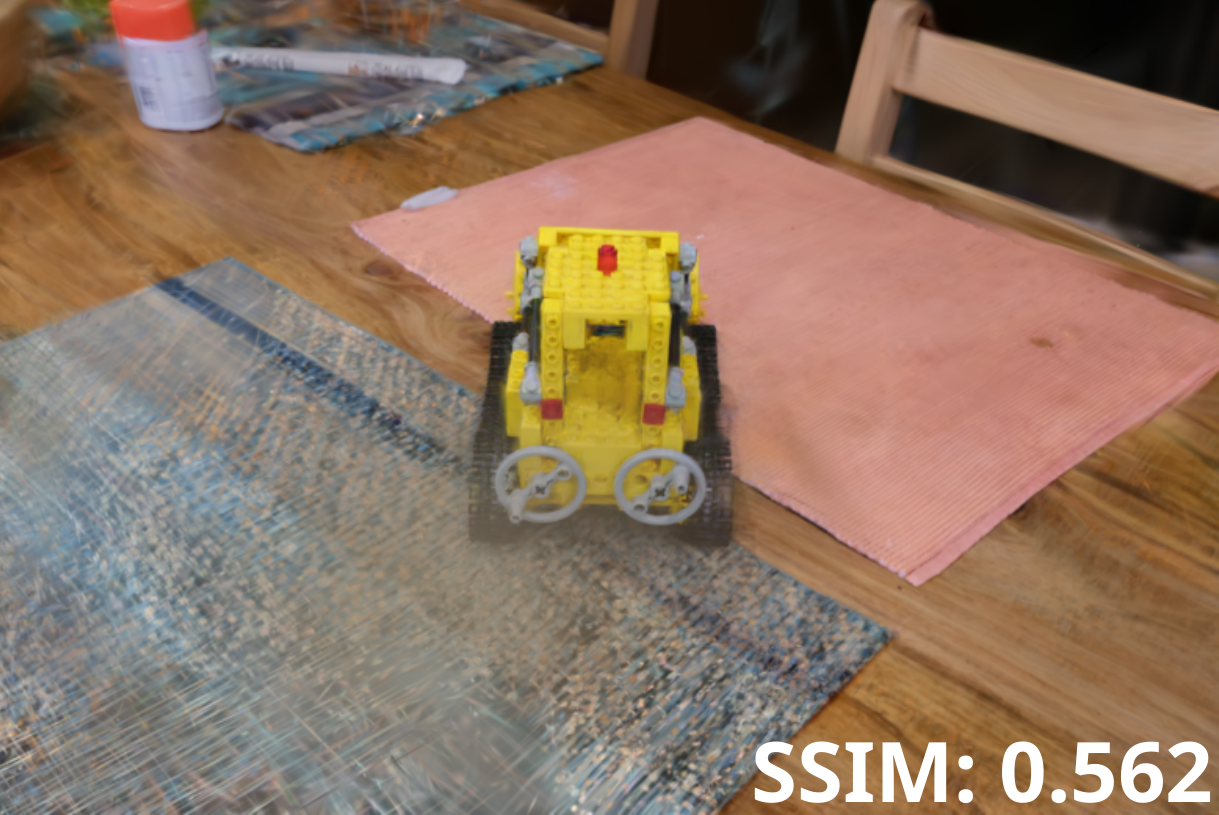}
    \caption*{\parbox[t]{\linewidth}{\centering\scriptsize + LPIPS Loss}}
    \label{fig:ablation_d}
  \end{subfigure}
  \hfill
  \begin{subfigure}[b]{0.32\columnwidth}
    \includegraphics[width=\textwidth]{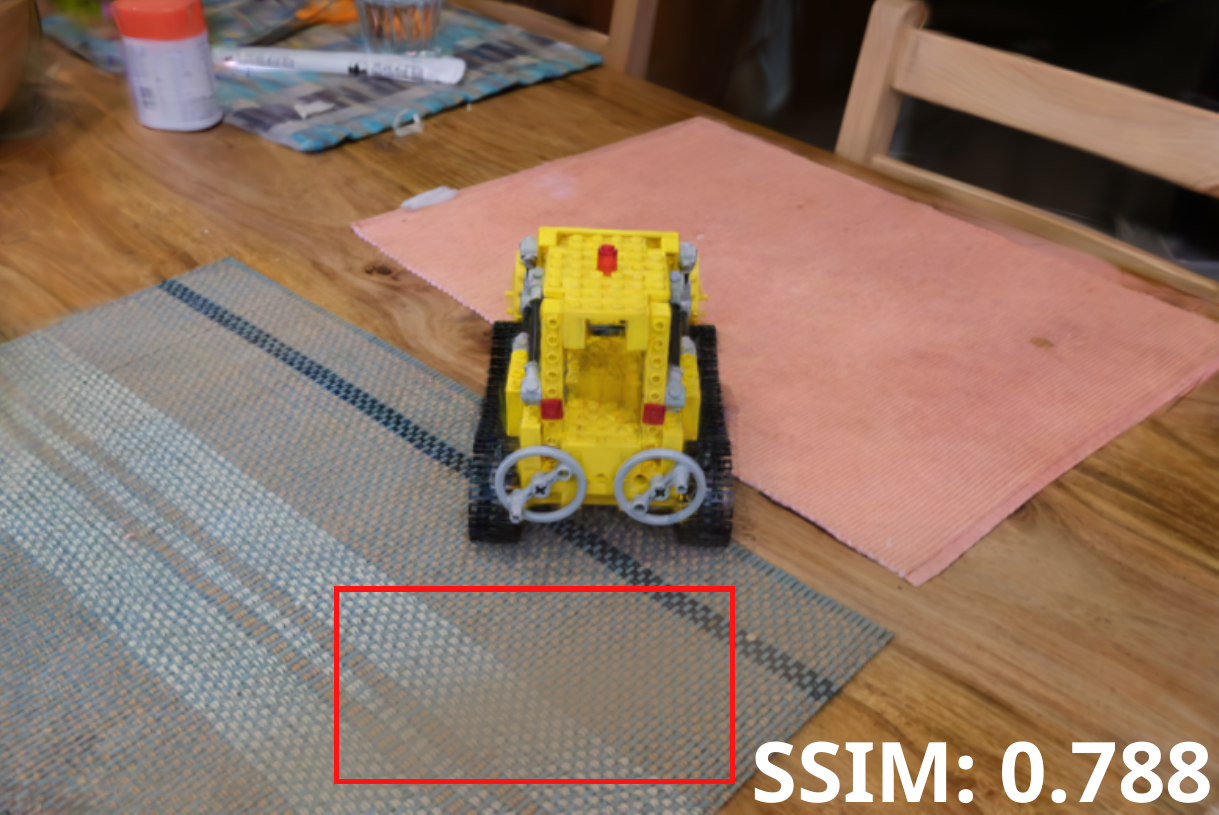}
    \caption*{\parbox[t]{\linewidth}{\centering\scriptsize + SfM \& Depth }}
    \label{fig:ablation_e}
  \end{subfigure}
  \hfill
  \begin{subfigure}[b]{0.32\columnwidth}
    \includegraphics[width=\textwidth]{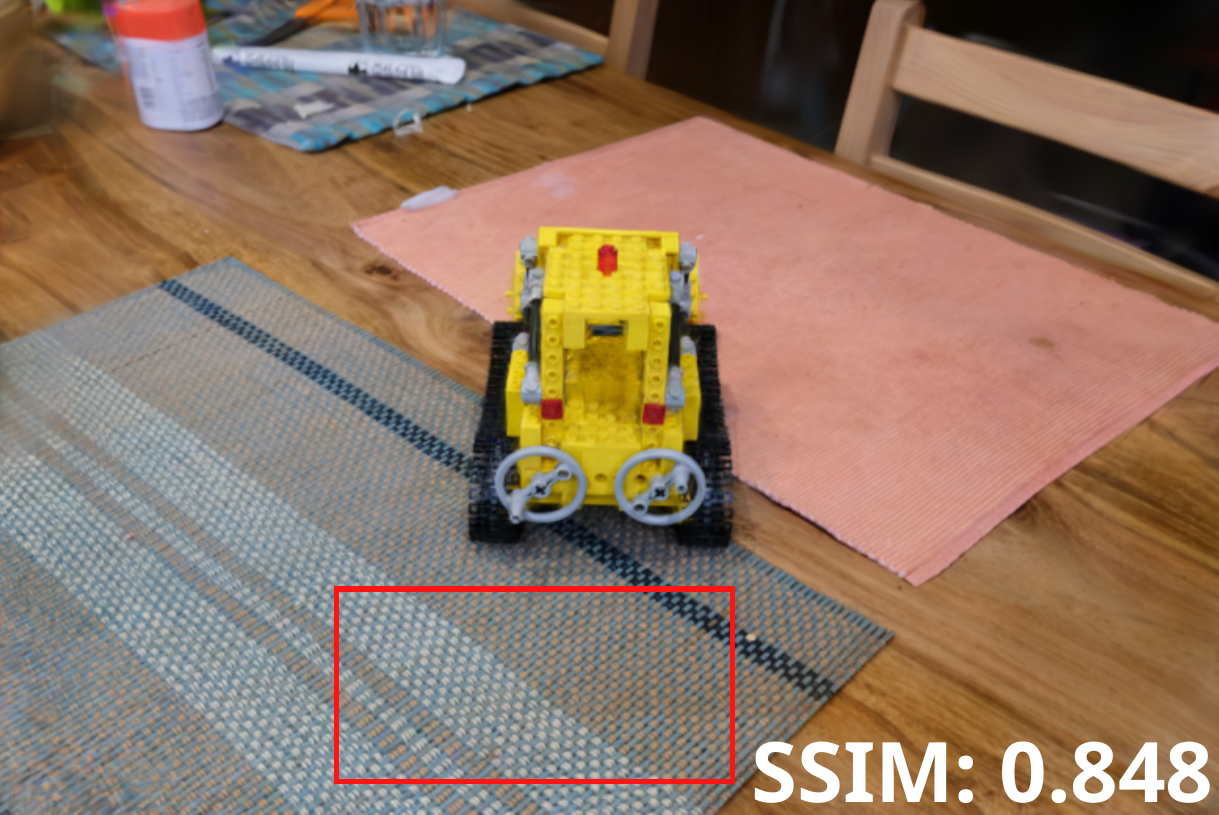}
    \caption*{\parbox[t]{\linewidth}{\centering\scriptsize \textbf{+Unc. Guide (Full)}}}
    \label{fig:ablation_f}
  \end{subfigure}

  \caption{
    \textbf{Qualitative Ablation Study on the Mip-NeRF 360 \cite{barron2022mipnerf360}} 'kitchen' scene for 24 views setting.} 
  \label{fig:ablation_qualitative}
\end{figure}

Figure~\ref{fig:ablation_qualitative} provides a visual breakdown of our ablation study. The baseline 3DGS model struggles with high-frequency texture on the woven mat. Naively incorporating synthetic views leads to oversmoothing and significant geometric errors, such as the distorted, widened shape of the Lego model. Our training schedule corrects this geometric corruption and oversmoothing, but results in similar distortions with the baseline, reducing the SSIM score. The addition of a perceptual LPIPS loss refines texture of the woven mat. While grounding the optimization with SfM initialization and a depth prior correctly scales the scene, it also introduces brown 'floater' artifacts on the mat, highlighted in the red rectangle. Finally, our final contribution, uncertainty-guided loss down-weighs inconsistent regions in the synthetic data, and removes the brown floaters on the mat, producing a clean, sharp final reconstruction.

\section{Conclusion}
\label{sec:conclusion}

In this work, we introduced OracleGS, a "propose-and-validate" framework that resolves the trade-off between generative completion and regressive fidelity in sparse-view novel view synthesis. Our method uses a pretrained 3D-aware diffusion model to propose complete scenes and repurposes an MVS model as an oracle to validate them. By distilling 3D uncertainty from the oracle's global-attention maps, we guide 3D Gaussian Splatting optimization to produce reconstructions that are both visually complete and geometrically sound, achieving state-of-the-art results on the Mip-NeRF 360~\cite{barron2022mipnerf360} and NeRF Synthetic~\cite{mildenhall2020nerf} benchmarks.

\paragraph{Acknowledgments.}
\label{sec:acknowledgments}
A. Murat Tekalp acknowledges support from Turkish Academy of Sciences (TUBA).

{
    \small
    \bibliographystyle{ieeenat_fullname}
    \bibliography{main}
}

\end{document}